\documentclass[10pt,twocolumn,letterpaper]{article}

\usepackage{cvpr}
\usepackage{times}
\usepackage{epsfig}
\usepackage{graphicx}
\usepackage{amsmath}
\usepackage{amssymb}
\usepackage{mathtools}

\cvprfinalcopy 

\def\cvprPaperID{15} 

\ifcvprfinal\fi
\begin{document}

\title{Synthesized Texture Quality Assessment via Multi-scale Spatial and Statistical
Texture Attributes of Image and Gradient Magnitude
Coefficients}

\author{S. Alireza Golestaneh \\
Arizona State University\\
{\tt\small sgolest1@asu.edu}
\and
Lina J. Karam\\
Arizona State University\\
{\tt\small karam@asu.edu}
}

\maketitle

\begin{abstract}
Perceptual quality assessment for synthesized textures is a challenging task.
In this paper, we propose a training-free reduced-reference  (RR) objective quality assessment method that quantifies the perceived quality of synthesized textures. 
The proposed reduced-reference synthesized texture quality assessment   metric is based on measuring the spatial and statistical attributes of the texture image using both image- and gradient-based wavelet coefficients at multiple scales.
Performance evaluations on two synthesized texture databases demonstrate that our proposed RR synthesized texture quality metric significantly outperforms both full-reference and RR state-of-the-art quality metrics in predicting the perceived visual quality of the synthesized textures\footnote{The source code of our proposed method will  be available online at  https://ivulab.asu.edu/software/IGSTQA/}.
\end{abstract}

\section{Introduction}
Natural and artificial textures  are important components in 	computer vision, image processing, multimedia, and graphics applications.
A visual texture consists of a spatially repetitive pattern of visual properties. 
Visual textures are present in both natural and man-made objects (e.g., grass, flowers, ripples of water, floor tiles, printed fabrics) and help in characterizing and recognizing these objects. 
A natural image typically consists of several types of visual texture regions that are present in the image, while a texture image corresponds to one such visual texture region.

\begin{figure}[t]
\centering
   \centerline{\includegraphics [scale=.46]{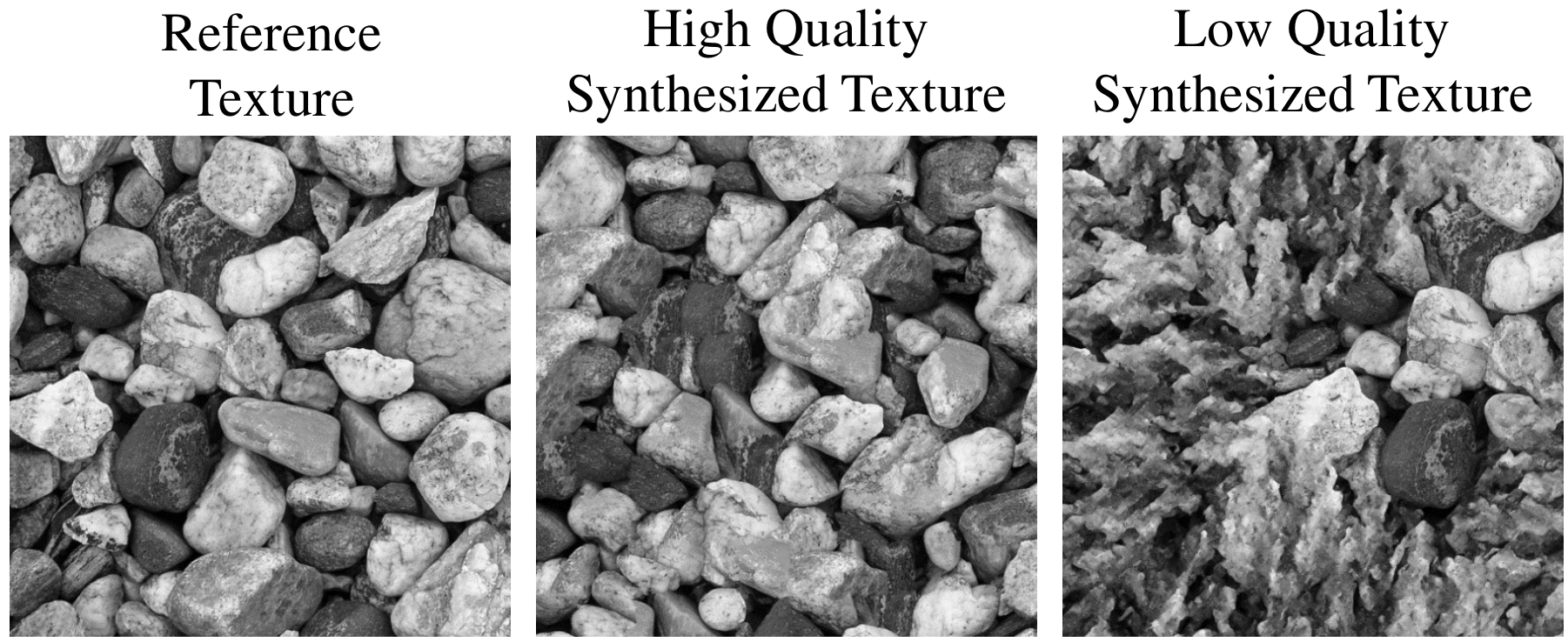}}
	\caption{Examples  of a reference texture as well as a high and low quality synthesized texture.}
	\label{Fig1}
 \end{figure}

\begin{figure}[t]
\centering
   \centerline{\includegraphics [scale=.42]{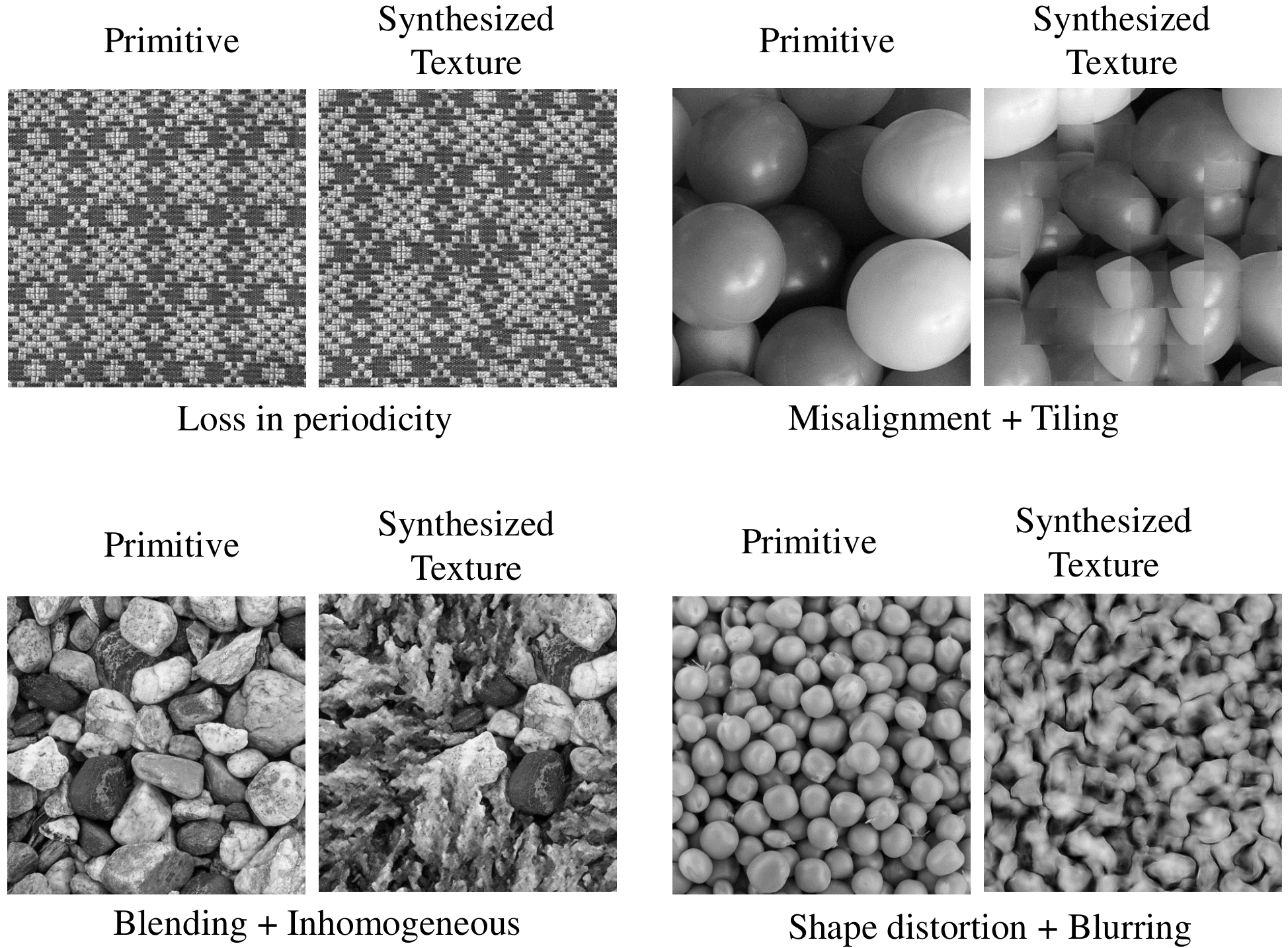}}
	\caption{Examples  of   reference textures (primitive)  as well as  their synthesized textures
	to illustrate the artifacts that can happen in texture synthesis.
	}
	\label{Fig22}
 \end{figure}

Texture synthesis is an important research topic; the use of an efficient synthesis algorithm can benefit many important applications in computer vision, multimedia, computer graphics, and image and video processing.
Applications of texture synthesis include image/video restoration \cite{wang2013efficient,kamel2008moving,yamauchi2003image}, image/video generation \cite{matsuyama2002generation,olszewski2017realistic,georgiadis2015texture}, image/video compression \cite{pappas2010image,balle2011models}, 
 multimedia image processing \cite{furht2012video, wolfgang1997overview},
 texture perception and description \cite{rao2012taxonomy,varadarajan2013no,
bergen1988early,keller1989texture,picard1995vision,
   lin2003finding,chetverikov2005brief,luo2013texture}, 
texture segmentation and recognition \cite{chaudhuri1995texture,manjunath2001color,salari1995texture,cimpoi2015deep, jaderberg2015spatial}, 
and synthesis \cite{efros2001image,portilla2000parametric,kwatra2003graphcut,ulyanov2016texture}.
Given a reference texture image and two corresponding synthesized versions, a human observer can easily determine which version better represents the original texture (see Figure \ref{Fig1}).
 However, automating this task is still very challenging  . 
Over the past
several decades, a large body of research has focused on developing accurate and efficient texture synthesis
algorithms.
Different texture synthesis methods produce different types of visual artifacts that lead to a loss in fidelity
of the synthesized textures compared to the original. 
These artifacts include misalignment, blur, tiling, and loss in the periodicity of the primitives (see Figure \ref{Fig22}).
The introduced artifacts alter the statistical properties in addition to the granularity and regularity attributes as compared to the original reference texture. 
Despite   advances in texture modeling and synthesis, there is little work  on developing algorithms for assessing the visual quality of   synthesized textures.

In natural image quality assessment, the assumption is that if a test image is high quality, the local structure of that test image should be very similar to the reference image. However, for synthesized texture quality assessment, the local structure of the synthesized texture may be different as compared to the reference texture, but the synthesized texture can still be perceived to be very similar to the reference texture if some main properties of the texture patterns are preserved. 
Therefore, in synthesized texture quality assessment, it is important to extract and quantify these texture attributes that convey the perceptually relevant information.  

The objective of   synthesized texture quality assessment is to provide computational models to measure the quality of a synthesized texture as perceived by human subjects.
 However, there are currently no satisfactory objective methods that can reliably estimate the perceived visual quality of   synthesized textures.
Based on the availability of a reference image,  objective quality metrics can be divided into full-reference (reference available or FR), no-reference (reference not available or NR), and reduced-reference (RR) methods.

FR methods usually provide the most precise evaluation results for natural images. However,   in many practical applications, the visual quality assessment (VQA) system does not have access to reference images.
RR visual quality assessment (RRVQA) methods provide a solution when the reference image is not completely accessible. 
These methods generally operate by extracting a   set of features from the reference image (RR features).
 The extracted RR features are later used with the distorted image (e.g., synthesized texture) to estimate quality. 
 RRVQA systems generally include a feature extraction process at the sender side for the reference image and a feature extraction at the receiver side for the distorted image. 
The RR features that are extracted from the reference image,  have a much lower data rate than the reference image data and are typically transmitted to the receiver through an ancillary channel \cite{wang2003objective}.

Given two synthesized versions of a visual texture, a human observer can easily select which of the two synthesized versions represents the reference texture better. 
However, this task is extremely challenging from a computational standpoint.
As it is shown later in this paper, existing modern objective VQA algorithms that are designed for natural images fail to accurately and reliably predict the quality of the synthesized textures.
 The process of automatically assessing the perceived visual quality of synthesized textures is ill-posed because of two reasons, namely,  (i) the sizes of the synthesized and the original texture
can be different and (ii)  the synthesized textures are not required to have pixel-wise correspondences with the original texture but can still appear perceptually equivalent (see Figure \ref{Fig1}).

Natural image statistics and structural similarity are used in existing popular objective image quality assessment (IQA) methods that are designed for natural images  \cite{wang2004image, wang2003multiscale,wang2005translation,moorthy2011blind, mittal2013making,zhang2015feature}.
SSIM \cite{wang2004image}  uses the mean, variance, and co-variance of pixels to compute luminance, contrast, and structural similarity, respectively.
MS-SSIM \cite{wang2003multiscale}  and CWSSIM \cite{wang2005translation} extended SSIM to the multiscale and  complex wavelet domain, respectively. 
State-of-the-art RR metrics such as RRIQA \cite{li2009reduced} and RRSSIM \cite{rehman2010reduced} require training and/or tuning of parameters to optimize the IQA performance.
Training-free RRIQAs \cite{xue2010reduced,mittal2012no,golestaneh2016reduced} usually need a large number of RR features  (side information) and their performance degrades with the reduction of the amount of side information.
 In \cite{mittal2012no}, Soundararajan et al. developed a training-free RRIQA framework (RRED) based on an information theoretic framework. 
The image quality is computed via the difference between the entropies of wavelet coefficients of reference and distorted images.
Golestaneh and Karam  \cite{golestaneh2015reduced}  proposed a training-free RRIQA based on the entropy of the divisive normalization transform of locally weighted gradient magnitudes. 
In \cite{xue2010reduced}, Xue et al. proposed a method ($\beta$W-SCM) based on the steerable pyramid.
The strongest component map (SCM) is constructed for each scale.
Then, the Weibull distribution is employed to describe the statistics of the SCM.
The Weibull scale parameters, one for each pyramid level, represent the RR features.

More specific to texture images, a   FR structural texture similarity index
(STSIM) was proposed in \cite{zhao2008structural,zujovic2013structural}.
Moreover,   Swamy et al. \cite{swamy2011parametric} proposed
an FR metric that uses Portilla's constraints \cite{portilla2000parametric} along
with the Kullback-Leibler Divergence (KLD). 
In \cite{varadarajan2014reduced}, Varadarajan and Karam proposed a RR VQA metric for texture synthesis based on visual attention and the perceived regularity of synthesized textures. 
The perceived regularity is quantified through the characteristics of the visual saliency map and its distribution.
Recently in \cite{golestaneh2016reduced}, Golestaneh and Karam proposed a training-free RR metric which is based on measuring the spatial and statistical texture attributes in the wavelet domain, namely granularity, regularity, and kurtosis.
 
In this paper, we propose a training-free RR Synthesized Texture   Quality Assessment method based on multi-scale spatial and statistical
texture attributes that are extracted from both image-based and gradient-based wavelet coefficients at different scales.   
We show as part of this work that higher performance can be obtained in terms of correlation with the perceived texture similarity by extracting multiscale statistical properties as well as regularity and granularity attributes from both the texture image and its gradient magnitude. 
In the proposed method, the RR features
are extracted from the considered texture image and its corresponding gradient magnitude image by first performing
an $L$-level multi-scale decomposition using an undecimated
wavelet transform. 
The perceived granularity and regularity of the texture image are
 quantified by computing, respectively, the mean and
the standard deviation of the locations of local extrema in wavelet
subbands based on the distribution of the  the absolute values of the wavelet coefficients in   each subband  \cite{golestaneh2016reduced}.
In addition, statistical RR features are extracted by computing the standard deviation, skewness, kurtosis, and entropy of the wavelet coefficient magnitude's distribution at each level of the multi-scale decomposition. 

The rest of this paper is organized as follows. Section 2
presents the proposed RR synthesized texture quality assessment
index. Performance results are presented in Section 3,
followed by a conclusion in Section 4.

\begin{figure}[t]
\centering
   \centerline{\includegraphics [scale=.38]{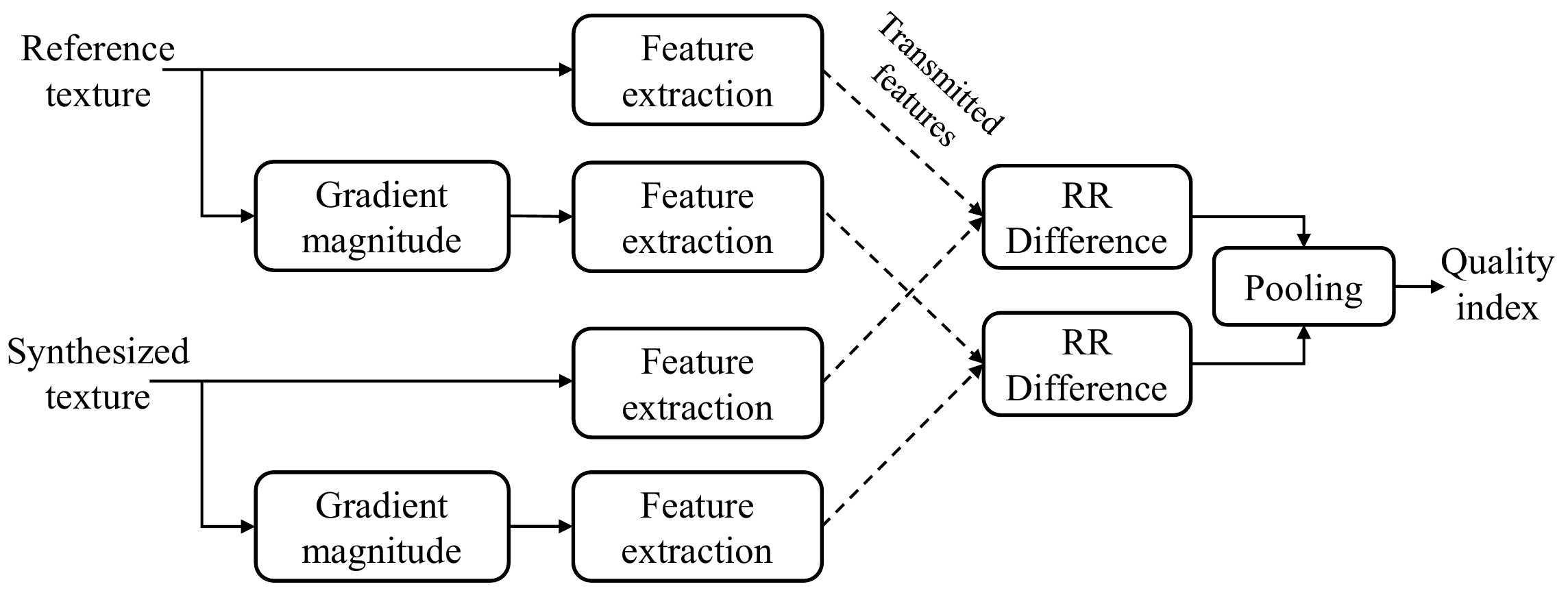}}
	\caption{ The general framework of our proposed IGSTQA method.}
	\label{Fig2}
 \end{figure}

\section{Proposed RR Visual Quality Assessment For Synthesized Textures}
Given an input reference texture and a   synthesized texture, Figure \ref{Fig2} shows    the framework of our proposed method.
In our proposed method, the RR features are extracted in the wavelet domain from both the spatial image $I$ and its gradient magnitude 
$I_{GM}$.
Perceptually relevant structures are further enhanced by combining properties from both the spatial and gradient magnitude domains. 
The image gradient is a popular feature in IQA \cite{xue2010reduced,zhang2011fsim,liu2012image}, since it can effectively capture local image structures, to which the HVS is highly sensitive. 
We compute the gradient magnitude $I_{GM}$ of the input image as the root mean square of the image directional gradients along two orthogonal directions.

\begin{figure}[t]
\centering
   \centerline{\includegraphics [scale=.53]{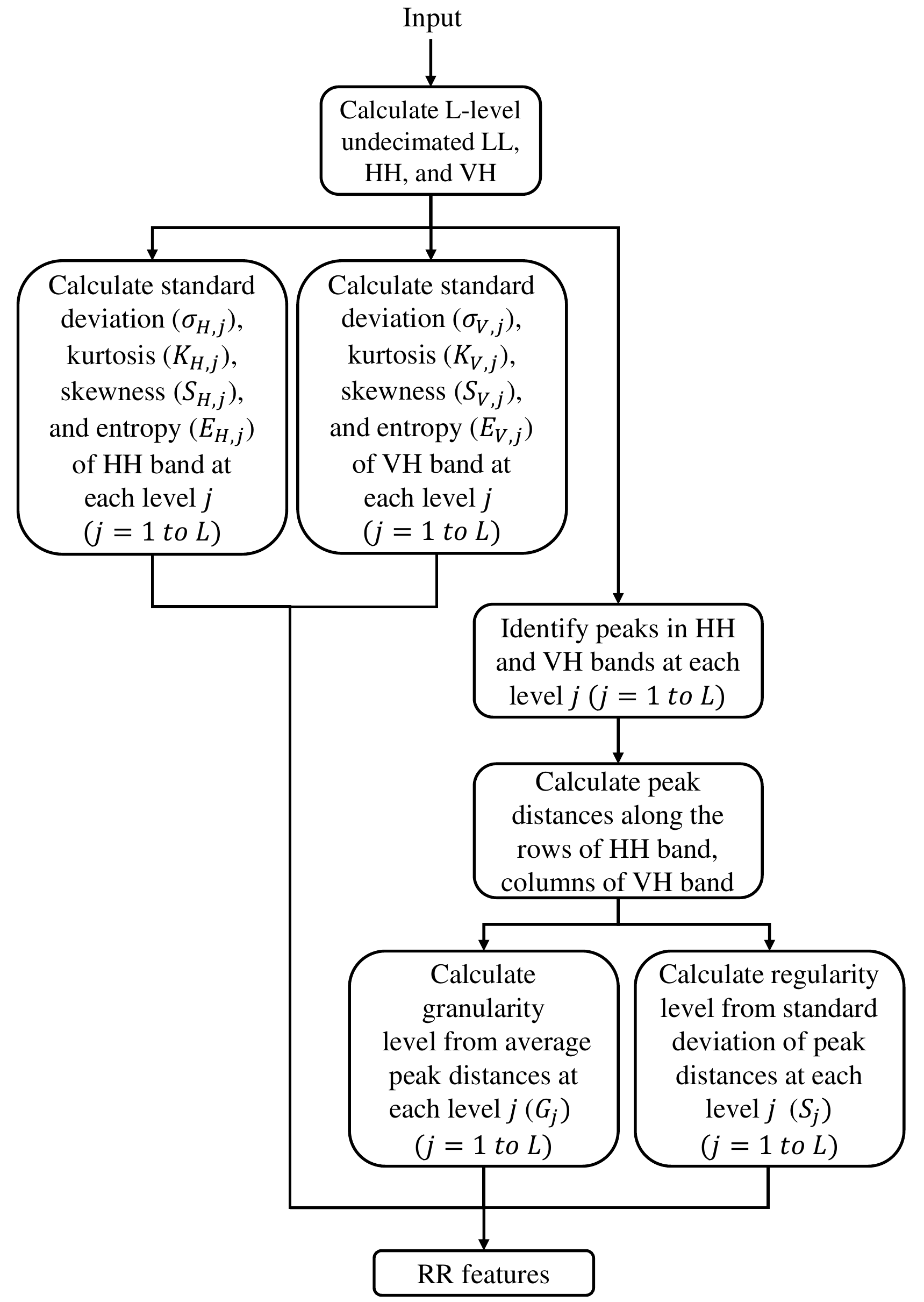}}
	\caption{ Block diagram illustrating the computation of the RR features for the proposed   index.}
	\label{Fig3}
 \end{figure}
 
The introduced artifacts while synthesizing the texture alter the statistical properties in addition to the granularity and regularity attributes as compared to the original reference texture. 
Different types of artifacts would be altering properties more significantly at a given scale and given domain and therefore we extract multiscale attributes and incorporate these in our proposed metric. For example, changes in attributes due to the blur artifact would be more pronounced in the high-frequency bands at lower scales and would affect the granularity attributes of the texture.
Tiling will also affect more significantly attribute in higher-frequency bands and would affect the regularity. 
Loss of periodicity of primitives affect significantly the regularity attribute. 

The wavelet-domain RR features are computed as shown
in Figure \ref{Fig3}.
First an $L$-level undecimated wavelet decomposition \cite{zhan2003wavelet} of the input texture image is performed ($L$ = 4 in our implementation), where the input  is divided at each level into three subbands, namely low-low (LL), horizontal-high (HH) and vertical-high (VH) subbands.
Our proposed quality index quantifies the perceived synthesized texture quality by extracting spatial features (granularity \cite{subedar2015no} and
regularity \cite{golestaneh2016reduced}) and statistical features (standard deviation, kurtosis, skewness, and entropy) at each scale.

The HH and VH subbands at the $j^{th}$        scale are denoted by $HH_j$ and $VH_j$, respectively. 
For
computing the granularity, $G_j$, and regularity, $R_j$, features at the $j^{th}$ scale, local peaks are detected by locating (as in \cite{subedar2015no})   the local maxima of the wavelet coefficients' magnitude  along the rows and columns of the $HH_j$ and $VH_j$ subbands, respectively.
Distances between adjacent located peaks are computed for every row (column) in the $HH_j$ ( $VH_j$ ) subband. 
Then the spatial features for the considered texture image, namely, the  granularity, $G_j$, and regularity, $R_j$, are computed, respectively, as the mean and standard deviation of the computed distances.

Let $G^M_{j,N}$ and $R^M_{j,N}$ denote the granularity and regularity  at the $j^{th}$ scale, where $M\in\{r,s\}$ and $N\in\{I, I_{GM}\}$,  with  $M = r $ denoting the reference texture,  $M= s$  denoting the synthesized texture.
 $N = I$  indicates that  the spatial image ($I$) is used to compute the RR features  while  $N = I_{GM}$ indicates   that the gradient magnitude image ($I_{GM}$) is used to compute the RR features.
Furthermore, let $\sigma^M_{H,j,N}$,  $K^M_{H,j,N}$, $S^M_{H,j,N}$,  $E^r_{H,j,N}$ denote, respectively, the  standard deviation,  kurtosis, skewness, and log energy entropy \cite{coifman1992entropy} of the $j^{th}$
level subband $HH_j$ corresponding to $M\in\{r,s\}$ and $N\in\{I,I_{GM}\}$. 
Similarly, let $\sigma^M_{V,j,N}$,  $K^M_{V,j,N}$, $S^M_{V,j,N}$,  $E^r_{V,j,N}$ denote, respectively, the  standard deviation,  kurtosis, skewness, and log energy entropy of the $j^{th}$   level subband $VH_j$. 
We define $\nabla K_{N}$ as:
\begin{equation}
\nabla K_{N}=\frac{\sum_{X\in\{H,V\}}\sum_{j=1}^{L}|K_{X,j,N}^{r}-K_{X,j,N}^{s}|}{2L},
\end{equation}
where $\nabla K_{N}$ denotes the distance between the kurtosis attributes $K_{X,j,N}^{r}$ and $K_{X,j,N}^{s}$ of the image ($N=I$) or 
gradient magnitude  ($N=I_{GM}$) wavelet coefficients' distribution.  
Similarly,  $\nabla \sigma_{N}$, $\nabla S_{N}$, and $\nabla E_{N}$ can be defined using Eq. (1) by replacing    $K$ with $\sigma$, $S$, and $E$, respectively.
Also, let $\nabla G_{N}$ denote the granularity difference between the original and synthesized texture in the raw image domain ($N=I$) or gradient magnitude image  domain ($N= I_{GM}$). $\nabla G_N$ can be defined as follows:
\begin{equation}
\begin{aligned}
\nabla G_{N} =
\frac{{\displaystyle \max_{j}}|G_{H,j,N}^{r}-G_{H,j,N}^{s}|}{2}+\\
\frac{{\displaystyle \max_{j}}|G_{V,j,N}^{r}-G_{V,j,N}^{s}|}{2},
\end{aligned}
\end{equation}
where $G_{H,j,N}^{r}$ ($G_{V,j,N}^{r}$) and $G_{H,j,N}^{s}$ ($G_{V,j,N}^{s}$) denote, respectively,  the granularity of the reference texture and the synthesized texture at the $j^{th}$ scale for the HH (VH)
subband.
Similarly, $\nabla R_{N}$ can be defined using Eq. (2) by replacing the granularity $G$ with the regularity $R$.

Finally, the proposed reduced-reference Image- and Gradient-based wavelet domain Syntheized Texture Quality Assessment (IGSTQA) index, is computed as follows:
\begin{equation}
\begin{gathered}
\scalebox{1.0}{$
\begin{aligned}
IGSTQA = \sum_{N\in\{I,I_{GM}\}}\log(1+\alpha(\nabla K_{N}+\nabla\sigma_{N}+\\
\nabla S_{N}+\nabla E_{N}+\nabla G_{N}+\nabla R_{N})).
\end{aligned}
$}
\end{gathered}
	\label{Eq1}
\end{equation}
In  Eq. (\ref{Eq1}), a value of $\alpha=100$   was found to yield  good results across a wide variety of images.
However, the selection of this value is not critical; the results
are very close when  $\alpha$  is chosen within a $\pm 20\%$ range.
\section{Results}
This section analyses the performance of our proposed method (IGSTQA) in terms of
qualitative and quantitative results.

\begin{figure}[t]
\centering
   \centerline{\includegraphics [scale=.26]{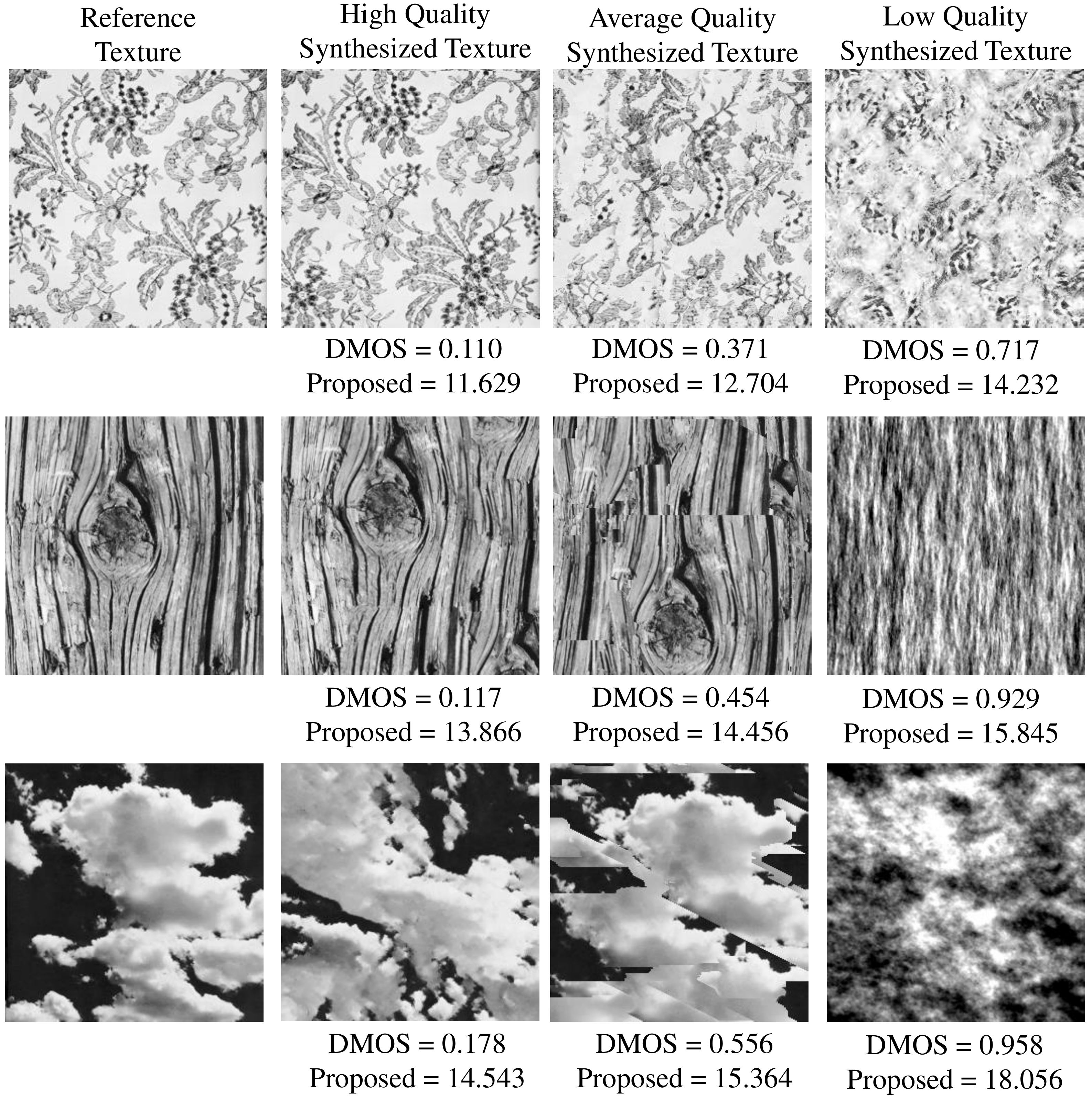}}
	\caption{Qualitative results for the proposed IGSTQA index for synthesized texture images taken from the Parametric Quality Assessment database \cite{swamy2011parametric}.}
	\label{Fig4}
 \end{figure}
 
\subsection{Qualitative Results}
Figure \ref{Fig4} provides results of our algorithm on three images with
different qualities.
 As shown in Figure  \ref{Fig4}, our algorithm can predict the
quality of 
texture 
images over a range of different qualities
in a manner that is consistent with human quality judgments
(DMOS).
 Notice that as we move from left to right within each
row, DMOS increases and IGSTQA follows a similar trend. In
terms of the across-image quality assessment, as we move from
top to bottom, DMOS increases and IGSTQA follows a similar
trend. 

\subsection{Quantitative Results}
In this section, the performance of the proposed IGSTQA
index is analyzed in terms of its ability to predict subjective
ratings of the synthesized texture quality. 
We evaluate the performance in
terms of prediction accuracy, prediction monotonicity, and
prediction consistency. 
To quantify the performance of our
algorithm, we applied IGSTQA to two different synthesized texture quality
databases including   the SynTEX Granularity \cite{golestaneh2015effect} and Parametric Quality Assessment \cite{swamy2011parametric} databases. 
 The SynTEX Granularity \cite{golestaneh2015effect} database  contains 21 reference  and 105 synthesized texture images that are generated by using five different texture synthesis algorithms, and the  Parametric Quality Assessment \cite{swamy2011parametric} database contains 42 reference textures and 252  synthesized texture images generated by using 6 different texture synthesis algorithms.


\begin{table}[t!]
\centering
\caption{Performance evaluation results of the proposed
IGSTQA index and comparison with IQA methods using the
SynTEX granularity database \cite{golestaneh2015effect}. Bold and italicized entries are
the best and second-best performers, respectively.} \label{tab2}
\resizebox{3.2in}{!} {
\begin{tabular}{c|c|r@{\extracolsep{0pt}.}lr@{\extracolsep{0pt}.}lr@{\extracolsep{0pt}.}l}
\multicolumn{1}{c}{} & \multicolumn{1}{c}{} & \multicolumn{2}{c}{} & \multicolumn{2}{c}{} & \multicolumn{2}{c}{}\tabularnewline
\hline 
\hline 
\multicolumn{8}{c}{SynTEX Granularity database \cite{golestaneh2015effect}}\tabularnewline
\multicolumn{1}{c}{} & \multicolumn{1}{c}{\# Features} & \multicolumn{2}{c}{PLCC} & \multicolumn{2}{c}{SROCC} & \multicolumn{2}{c}{RMSE}\tabularnewline
\hline 
\hline 
PSNR & FR & 0&237  & 0&345 &  1&210\tabularnewline
MS-SSIM \cite{wang2003multiscale} & FR & 0&293 & 0&122 & 1&105 \tabularnewline
STSSIM \cite{zujovic2013structural} & FR & 0&215 & 0&135 & 1&213\tabularnewline
CWSSIM \cite{wang2005translation} & FR & 0&595 & 0&583  & 0&914\tabularnewline
Parametric \cite{swamy2011parametric} & FR & 0&487  & 0&328 & 1&087\tabularnewline
DIIVINE \cite{moorthy2011blind} & NR & 0&357 & 0&408 & 1&094\tabularnewline
NIQE\cite{mittal2013making} & NR & 0&253 & 0&218 & 1&154\tabularnewline
IL-NIQE \cite{zhang2015feature} & NR & 0&543 & 0&512 & 0&985\tabularnewline
RRED \cite{soundararajan2012rred} & $\frac{Image\:Size}{32}$ & 0&226  & 0&116 & 1&211\tabularnewline
$\beta$W-SCM \cite{xue2010reduced} & 6 & 0&472 & 0&415 & 1&158\tabularnewline
STQA  \cite{golestaneh2016reduced} & 7 & \emph{0}&\emph{770} & \emph{0}&\emph{777} & \emph{0}&\emph{792}\tabularnewline
Proposed & $24L$ & \textbf{0}&\textbf{816} & \textbf{0}&\textbf{820} & \textbf{0}&\textbf{718}\tabularnewline
\hline 
\hline 
\multicolumn{1}{c}{} & \multicolumn{1}{c}{} & \multicolumn{2}{c}{} & \multicolumn{2}{c}{} & \multicolumn{2}{c}{}\tabularnewline
\end{tabular}}
\end{table}

 \begin{table}[]
\caption{Performance evaluation results of the proposed
IGSTQA index and comparison with IQA methods using the Parametric
Quality Assessment database  \cite{swamy2011parametric}. Bold and italicized entries
are the best and second-best performers, respectively.} \label{tab1}
\resizebox{3.2in}{!} {
\begin{tabular}
{c|c|r@{\extracolsep{0pt}.}lr@{\extracolsep{0pt}.}lr@{\extracolsep{0pt}.}l}
\multicolumn{1}{c}{} & \multicolumn{1}{c}{} & \multicolumn{2}{c}{} & \multicolumn{2}{c}{} & \multicolumn{2}{c}{}\tabularnewline
\hline 
\hline 
\multicolumn{8}{c}{Parametric Quality Assessment database \cite{swamy2011parametric}}\tabularnewline
\multicolumn{1}{c}{} & \multicolumn{1}{c}{\# Features} & \multicolumn{2}{c}{PLCC} & \multicolumn{2}{c}{SROCC} & \multicolumn{2}{c}{RMSE}\tabularnewline
\hline 
\hline 
PSNR & FR & 0&083 & 0&075 & 0&952\tabularnewline
MS-SSIM \cite{wang2003multiscale} & FR & 0&087  & 0&053 &  0&921\tabularnewline
STSSIM \cite{zujovic2013structural} & FR & 0&045 & 0&054 & 0&964\tabularnewline
CWSSIM \cite{wang2005translation} & FR & 0&015  & 0&002 &  0&953\tabularnewline
Parametric \cite{swamy2011parametric} & FR & 0&412 & 0&481 & 0&253\tabularnewline
DIIVINE \cite{moorthy2011blind} & NR & 0&351 & 0&203 & 0&254\tabularnewline
NIQE \cite{mittal2013making} & NR & 0&185 & 0&054 & 0&315\tabularnewline
IL-NIQE \cite{zhang2015feature} & NR & 0&432 & 0&403 & 0&253\tabularnewline
RRED \cite{soundararajan2012rred} & $\frac{Image\:Size}{32}$ & 0&208  & 0&188 & 0&255\tabularnewline
$\beta$W-SCM \cite{xue2010reduced} & 6 & 0&375 & 0&398 & 0&254\tabularnewline
STQA \cite{golestaneh2016reduced} & 7 & \emph{0}&\emph{532} & \emph{0}&\emph{520} & \emph{0}&\emph{250}\tabularnewline
Proposed & $24L$ & \textbf{0}&\textbf{733 } & \textbf{0}&\textbf{679} & \textbf{0}&\textbf{170}\tabularnewline
\hline 
\hline 
\multicolumn{1}{c}{} & \multicolumn{1}{c}{} & \multicolumn{2}{c}{} & \multicolumn{2}{c}{} & \multicolumn{2}{c}{}\tabularnewline
\end{tabular}}
\end{table}

We employ three commonly used performance
metrics. We measure the prediction monotonicity of
IGSTQA via the Spearman rank-order correlation coefficient
(SROCC). We measure the Pearson linear correlation
coefficient (PLCC) between MOS (DMOS) and the objective
scores after nonlinear regression. The root mean squared error
(RMSE) between MOS (DMOS) and the objective scores
after nonlinear regression is also measured.
Tables 1 and 2 provide the comparison between our results and popular FR, RR, and NR IQA algorithms  using the SynTEX granularity database  \cite{golestaneh2015effect} and the Parametric Quality Assessment \cite{swamy2011parametric} databases, respectively.
 The results show that the modern FR, NR,  and RR metrics do not perform well for quantifying the quality of synthesized textures.
 Moreover, it can  be observed that our proposed  quality index
yields the highest correlation with the subjective quality ratings in terms of PLCC, SROCC, and RMSE.

Table 3  shows the performance of our proposed algorithm  in terms of PLCC,
SROCC, and RMSE when either the spatial or gradient magnitude domain is used to extract the wavelet coefficients. 
From Table 3 it can be seen that extracting attributes from only one domain only decreases the performance of the proposed method, as compared to incorporating     attributes from both domains.

%
 
\begin{table}[h]
\caption{ Performance evaluation of IGSTQA while using just the spatial or gradient magnitude domains.} \label{tab3}
\resizebox{3.3in}{!} {
\begin{tabular}{c|c|r@{\extracolsep{0pt}.}lr@{\extracolsep{0pt}.}lr@{\extracolsep{0pt}.}l}
\multicolumn{1}{c}{} & \multicolumn{1}{c}{} & \multicolumn{2}{c}{} & \multicolumn{2}{c}{} & \multicolumn{2}{c}{}\tabularnewline
\hline 
\multicolumn{1}{c}{Database} & \multicolumn{1}{c}{Criterion} & \multicolumn{2}{c}{PLCC} & \multicolumn{2}{c}{SROCC} & \multicolumn{2}{c}{RMSE}\tabularnewline
\hline 
\hline 
Parametric Quality  & Using spatial domain & 0&682 & 0&651 & 0&213\tabularnewline
Assessment database & Using gradient magnitude domain & 0&702 & 0&664 & 0&197\tabularnewline
 \cite{swamy2011parametric} & Proposed & 0&733  & 0&679 & 0&170\tabularnewline
\hline 
SynTEX Granularity & Using spatial domain & 0&785 & 0&776 & 0&854\tabularnewline
database & Using gradient magnitude domain & 0&797 & 0&809 & 0&753\tabularnewline
 \cite{golestaneh2015effect} & Proposed & 0&816 & 0&820 & 0&718\tabularnewline
\hline 
\multicolumn{1}{c}{} & \multicolumn{1}{c}{} & \multicolumn{2}{c}{} & \multicolumn{2}{c}{} & \multicolumn{2}{c}{}\tabularnewline
\end{tabular}}
\end{table}
 
\section{Conclusion}
Finding a balance between the
number of RR features and the predicted image quality is
at the core of the design of RR VQA methods. 
Moreover, estimating the quality of synthesized textures is a very challenging
task.
In this paper,
we proposed an RR-training-free VQA method to assess the perceived visual quality
of synthesized textures based on spatial and statistical features extracted from the wavelet transform of both the texture image
and  its gradient magnitudes.
Our proposed RR index, IGSTQA, yields   the highest  
prediction accuracy for measuring the perceived fidelity of
synthesized textures and outperforms    state-of-the-art
  quality metrics.

\end{document}